# A Learning Algorithm for Evolving Cascade Neural Networks

Vitaly Schetinin
TheorieLabor, Friedrich-Schiller University of Jena
Ernst-Abbe-Platz 4, 07740 Jena, Germany
http://nnlab.tripod.com

**Abstract.** A new learning algorithm for Evolving Cascade Neural Networks (ECNNs) is described. An ECNN starts to learn with one input node and then adding new inputs as well as new hidden neurons evolves it. The trained ECNN has a nearly minimal number of input and hidden neurons as well as connections. The algorithm was successfully applied to classify artifacts and normal segments in clinical electroencephalograms (EEGs). The EEG segments were visually labeled by EEG-viewer. The trained ECNN has correctly classified 96.69% of the testing segments. It is slightly better than a standard fully connected neural network.
**Key words:** neural network, cascade architecture, evolving, feature selection, electroencephalogram

## 1. Introduction

To build feed-forward neural networks, a cascade-correlation learning algorithm [1] has been suggested which creates the hidden neurons as they are needed. However the algorithm applied to real-world problems often over-fit cascade networks and their testing error becomes to be more than the training error [2]. As we know, over-fitting may occur if the training data are characterized by many irrelevant and noisy features.

To overcome this problem different techniques of pre-processing the data have been developed, which are able to select the informative features. For example, one may use the ID3 and other feature selection algorithms [3, 4]. However, the results of these classification algorithms depend on some special conditions, for example, on the order in which the features are processed.

To prevent the cascade neural networks from over-fitting, in [5] a method based on a combination of two algorithms, early stopping and ensemble averaging, was developed. The authors showed that their method improves the prediction ability of neural networks. They also proposed an algorithm to estimate the generalization ability of their method using the leave-one-out technique.

The pruning methods described in [6] have been developed for networks trained by the cascade-correlation learning algorithm. These methods were used to estimate the importance of large sets of initial variables that characterize quantitative relationships. The results calculated by cascade-correlation networks were compared with the performance of fixed-size neural networks. The developed methods were successfully used to optimize the set of initial variables. The use of variables, the developed methods selected, results in an improvement of the prediction ability of the neural network.

To avoid over-fitting, a Group Method of Data Handling (GMDH) [7] has been suggested which allows to generate neural networks of appropriate complexity. To learn networks to generalize well, the GMDH exploits a regularity criterion calculated on the training and validating examples. The GMDH-type neural networks were successfully applied to real world problems [8].

In this paper we describe a learning algorithm that is able to select informative features while the cascade network is learning. We guess that is the most effective way to prevent the networks from over-fitting. In contrast to fully connected cascade networks, in this case the cascade network starts to learn with a small number of inputs. The network uses the new fea-tures during learning and for this reason we call such networks Evolving Cascade Neural Networks (ECNN).

The training algorithm has applied to classify the normal segments and artifacts in clinical electroencephalograms (EEGs). The features calculated to characterize the EEG segments may be irrelevant, e. g., noisy and redundant [9, 10, 11]. The EEG recordings we used were taken from two patients. All segments were visually labeled by an EEG-expert. The ECNN that learned to recognize the EEG artifacts has correctly classified 96.69% of the testing segments.

In Section 2 we will introduce and describe ECNNs, then in Section 3 and 4 describe in detail a fitting and ECNN training algorithms we developed. In Section 5, we will apply our algorithm to the real-world problem of EEG cleaning. Then in section 6 we will compare an ECNN and a standard feed-forward neural network technique on these EEG data. Finally in Section 7 and 8 we will discuss our results and effectiveness of our algorithm developed to train ECNNs.

## 2. Evolving Cascade Neural Networks

Let us define a cascade network architecture consisting of neurons, whose number $p$ of inputs is increased from one layer to the next. The neuron of the first layer is connected to two input nodes from the $x_1, \ldots, x_m$. The first of these inputs provides a minimal single neuron error. The neuron of the next layer is linked to the first input and to the other one as well as to the output of the previous neuron. Thus, the $r$-th neuron is connected to two input nodes and also to the outputs of all previous neurons.

Following this architecture, we can describe the output $z_r$ of the $r$-th neuron with $p = r + 1$ inputs as follows

$$z_r = f(\mathbf{u}, \mathbf{w}) = 1/(1 + \exp(-\Sigma_i^p u_i w_i)), \tag{1}$$

where $r = 1, 2, \ldots$ is the number of layer, $f$ is an activation function, $\mathbf{u} = (u_1, \ldots, u_p)$ is a $p \times 1$ input vector of the $r$-th neuron, and $w_0, \ldots, w_p$ are the components of a weight vector $\mathbf{w}$.

As the irrelevant features cause over-fitting of the neural networks, we can experimentally estimate their significance in an ad hoc, trial-and-error, manner. Accordingly, the algorithm, which we will outline in Section 4, starts to train the output neuron with one input, and then, step-by-step, add the new inputs as well as the new hidden neurons.

Therefore, the algorithm builds internal connections of the cascade neural network according to the values of a predefined fitness function. This fitness function could be defined as the regularity criterion used in the GMDH.

The regularity criterion denoted by $C_r$ is calculated for each neuron on the unseen examples, which were not used for fitting the connection weights. In this case the value of the criterion depends on the generalization ability of the trained neuron with the given connections: the value of the criterion is increased if the number of the misclassified examples increases. In other words, the neuron with irrelevant connections is not able to properly classify the unseen examples, and hence the value of the criterion is will be higher.

The idea behind our algorithm is to use the regularity criterion for selecting the neurons with relevant connections. According to this idea, if the value $C_r$ calculated for the $r$-th neuron is less than a value $C_{r-1}$ calculated for the previous, $(r - 1)$, neuron, then the features that feed the $r$-th neuron are relevant, else they are irrelevant. This can be described by the following inequality:

*if* $C_r < C_{r-1}$, \hfill (2)
*then* the features are relevant;

*else* they are irrelevant.

If inequality (2) is met, then the connections and the weights of the *r*-th neuron are saved, and this neuron is added to the network. In the case, where no neuron satisfies this inequality, the algorithm is stopped. As a result, a neuron with minimal value of a criterion is assigned to be the output one.

Note that due to noise in training data, the inequality (2) may be met for the (*r* + 1)-th neuron, but not for the *r*-th neuron. In this case a trained network is nearly optimal. However we can easily increase chance of finding out a proper network by increasing the number of unseen examples as well as the number of training runs. In our experiments described in Section 5 we have kept a half of the training data as the unseen examples and run the ECNN algorithm 100 times.

## 3. Fitting of Weights

As we have no a priori information about the noise in the training dataset, we will use a projection method described in [8] which allows to effectively fit the neuron weights in the presence of the unknown noise.

To implement the regularity criterion, a training dataset must be divided into at least two subsets, say A and B. One subset is used for fitting the neuron weights and the other one for validating this neuron. In this case, the output of the neuron with irrelevant connections calculated on the validating dataset is considerably different from desirable values. Thus, calculating a residual error on the validating dataset, we can control the training of neurons and a network.

Let $\mathbf{D} = (\mathbf{X}, \mathbf{Y}^0)$ be a training dataset consisting of *n* examples, where $\mathbf{X}$ is a $n \times m$-matrix of input data, and $\mathbf{Y}^0 = (y_1^0, ..., y_n^0)$ is a $n \times 1$ target vector. Let us divide a set $\mathbf{D}$ into two non-intersecting subsets $\mathbf{D}_A = (\mathbf{X}_A, \mathbf{Y}_A^0)$ and $\mathbf{D}_B = (\mathbf{X}_B, \mathbf{Y}_B^0)$, $n = n_A + n_B$, where $n_A$ and $n_B$ are the number of the examples in these subsets, respectively. Note that users may set $n_A$ and $n_B$, e.g., as $n_A \approx n_B$. The subsets $\mathbf{D}_A$ and $\mathbf{D}_B$ we will use for fitting and validating the weights, respectively.

In according to the ECNN structure, the input $\mathbf{u}_1$ of the first neuron is given by two features

$\mathbf{u}_1 = (x_i, x_{j1}), i \neq j_1 = 1,..., m.$

For the second neuron, the input is given by

$\mathbf{u}_2 = (z_1, x_i, x_{j2}), i \neq j_2 = 1,..., m,$

where $z_1$ is the output of the first neuron.

Then, for the *r*-th neuron, we have

$\mathbf{u}_r = (z_{r-1},..., z_1, x_i, x_{jr}), i \neq j_r = 1,..., m,$

where $z_i$ is the output of the *i*-th hidden neuron.

Since the training and validating of the neuron are realized on different subsets $\mathbf{D}_A$ and $\mathbf{D}_B$, let us denote its input data by $\mathbf{u}_A$ and $\mathbf{u}_B$, respectively, both are the $p \times 1$ vectors. Correspondingly, the *i*-th example from $\mathbf{D}$ we denote as $\mathbf{u}^{(i)} \in \mathbf{X}$, $\mathbf{u}^{(i)} = (u_1^{(i)}, ..., u_p^{(i)})$, and $y_i^0 \in \mathbf{Y}^0$. Using these notations, we can describe the basic steps of the fitting algorithm.

Initially, $k$ is set to zero, and an algorithm initiates a weight vector $\mathbf{w}^0$ by random values that are, e.g., Gaussian distributed. Then, at step $k = 1, 2, \ldots$, the algorithm calculates the $n_A \times 1$ error vector $\boldsymbol{\eta}_A^k$ on the training dataset $\mathbf{D}_A$. Its elements $\eta_{Ai}^k$, $i = 1, \ldots, n_A$, are calculated as

$$\eta_{Ai}^k = f(\mathbf{u}_A, \mathbf{w}^{k-1}) - y_A^0, \text{ where } y_A^0 \in \mathbf{Y}_A^0. \tag{3}$$

Correspondingly to a dataset $\mathbf{D}_B$, it is calculated the elements $\eta_{Bi}^k$, $i = 1, \ldots, n_B$, of the $n_B \times 1$ error vector $\boldsymbol{\eta}_B^k$:

$$\eta_{Bi}^k = f(\mathbf{u}_B, \mathbf{w}^{k-1}) - y_B^0, \text{ where } y_B^0 \in \mathbf{Y}_B^0.$$

Using the elements of the vector $\boldsymbol{\eta}_B^k$, we can then calculate an error $e_B$ of the neuron on the validating dataset:

$$e_B(k) = (\Sigma_i \eta_{Bi}^k)^{1/2}, i = 1, \ldots, n_B. \tag{4}$$

Note, that the residual error $e_B$ corresponds to the fitness function $C_r$ of the neuron, i.e., $C_r = e_B$.

We do not know the level of noise in the training dataset. However we can preset a constant $\Delta > 0$ which defines a minimal decrement of the error $e_B$, calculated at the step $(k-1)$ and $k$, respectively. Then the goal of the training algorithm is achieved, if the error difference between step $k^*$ and $(k^*-1)$ will be less than $\Delta$:

$$e_B(k^*-1) - e_B(k^*) < \Delta. \tag{5}$$

If this inequality is not satisfied, then the current weight vector $\mathbf{w}^{k-1}$ is updated in accordance to the following learning rule

$$\mathbf{w}^k = \mathbf{w}^{k-1} - \chi \| \mathbf{U}_A \|^{-2} \mathbf{U}_A \boldsymbol{\eta}_A^{k-1}, \tag{6}$$

where $\chi$ is the learning rate, $\mathbf{U}_A$ is the $p \times n_A$ matrix of the input data, $\|\mathbf{U}_A\| = (\Sigma_i^p u_{Ai}^{(1)2} + \Sigma_i^p u_{Ai}^{(2)2} + \ldots)^{1/2}$ is an Euclidian norm of $\mathbf{U}_A$.

Thus, after $k^*$ steps, the algorithm provides a desired weight vector $\mathbf{w}^*$. In our experiments the best performance was obtained with $\chi = 1.9$ and $\Delta = 0.0015$. In this case the number $k^*$ usually did not exceed 30 steps.

## 4. The ECNN Training Algorithm

The first heuristic we used is to exploit the best feature that provides a minimal value of *CR* calculated for the neurons with one input node. This heuristic is performed to find the first input $x_{i1}$.

The second heuristic is to connect an input node $x_{i1}$ to the candidate-neuron. This heuristic is applied to each candidate-neuron.

The third heuristic is expressed by inequality (2) that allows to find out those features that reduce current value of *CR*. This heuristic is also applied to each candidate-neuron.

Using these heuristics we developed the ECCN training algorithm main steps of which are as follow:

1. Initialize the layer $r = 0$ and a set $X := (x_1, \ldots, x_m)$. Calculate the values $S_i$ of fitness function $CR$ for neurons with one input node $x_i$, $i = 1, \ldots, m$.

2. Arrange the calculated values $S_i$ in ascending order and put them in a list $S$: $S := \{S_{i1} \leq S_{i2} \leq \ldots \leq S_{im}\}$. Set a value $C_0 = S_{i1}$.

3. Set a position $h = 2$ for the next feature in a set $X$ and a list $S$.

4. Set $r := r + 1$ and $p = r + 1$. Create the new candidate-neuron with $p$ inputs.

5. If $r > 1$, then connect the first $r$ inputs of this neuron to all previous neurons and to an input $x_{i1}$, respectively. Otherwise, connect this neuron to an input $x_{i1}$.

6. Connect a $p$-th input of a candidate-neuron to an input node being in a position $h$ of a set $X$.

7. Train the candidate-neuron and then calculate its value $C_r$.

8. If $C_r \geq C_{r-1}$, then go to step 10.

9. Put the candidate-neuron to the network as the $r$-th neuron. Go to step 4.

10. If $h < m$, then $h := h + 1$ and go to step 6, else stop.

In this algorithm, the weights of candidate-neurons are updated by formula (6) as long as condition (5) is not satisfied. At step 1, the neurons start to learn with one input. At the following steps, the growing network involves new features as well as new neurons while the value of error (4) is decreasing.

Finally, the trained network consists of a near minimal number of connections and neurons. Such a network, as we know, is able to generalize well.

Below, we will apply this algorithm to recognize the artifacts in the clinical EEGs. Note that in this real-world task the EEG data are characterized by many irrelevant features.

**5. Application to Electroencephalograms**

We used clinical EEG recordings made via two standard electrodes C3 and C4. Following [11], each EEG segment was represented by 72 calculated features related to the power spectral densities, calculated on 10-second segments into 6 frequency bands: sub-delta (0-1.5 Hz), delta (1.5-3.5 Hz), theta (3.5-7.5 Hz), alpha (7.5-13.5 Hz), beta 1 (13.5-19.5 Hz), and beta 2 (19.5-25 Hz). The power densities were calculated for C3 and C4 electrodes as well as for their total sum. For these 18 original features, the relative and absolute power values as well as their variances were calculated. The EEG data were finally normalized.

In our experiment, the EEG recordings have been made from two patients. The artifacts in both recordings were manually labeled by an EEG-viewer. These recordings were merged in a dataset that was divided into the training and testing subsets consisting of 2244 and 1210 randomly selected segments including the 209 and 99 artifacts, respectively. The subsets $\mathbf{D}_A$ and $\mathbf{D}_B$ have been composed of 1122 odd and even training examples, respectively.

Since the initial weights of the ECNN were randomly assigned, we have run the training algorithm 100 times. Then we selected a trained ECNN, whose error rate on the training set was minimal. In our experiment it was equal to 3.92% of the 2244 training examples. On the testing set consisted of the 1210 examples an error rate of this network was equal to 3.31%.

Fig. 4 depicts a structure of this network, containing four input nodes, three hidden and one output neurons. The training algorithm has selected from 72 only four features $x_{36}$, $x_{23}$, $x_{10}$, and $x_{60}$, which input to a network. The inputs of the first hidden neuron are connected to inputs $x_{36}$ and $x_{23}$. The inputs of the output neuron are connected to the outputs $z_1$, $z_2$ and $z_3$ of the hidden neurons and to the inputs $x_{36}$ and $x_{60}$.

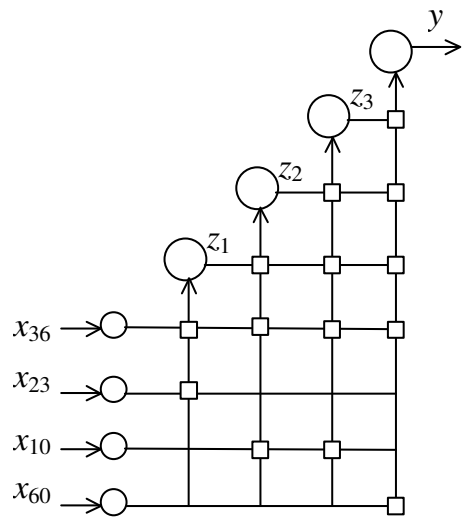

*Figure 1.* The structure of an ECNN trained to recognize the EEG artifacts consists of three hidden neuron and one output neuron.

During 100 runs, the sizes of the trained ECNNs varied from one neuron to 12. Note that the ECNN consisting of four neurons appears in most cases. A distribution of the network sizes is depicted in the histogram Fig. 2.

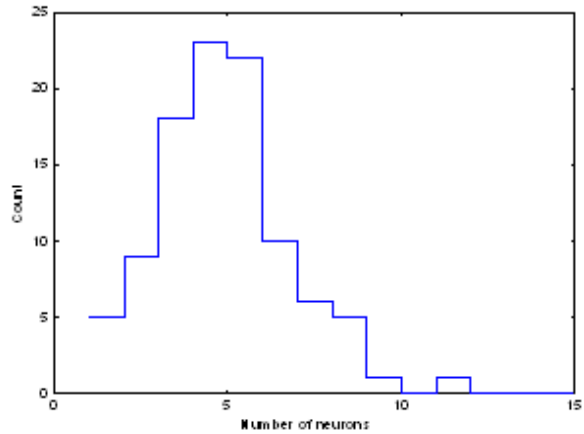

*Figure 2*. A histogram of the network sizes on 100 runs. The number of neurons in the trained ECNN was varied from 1 to 12.

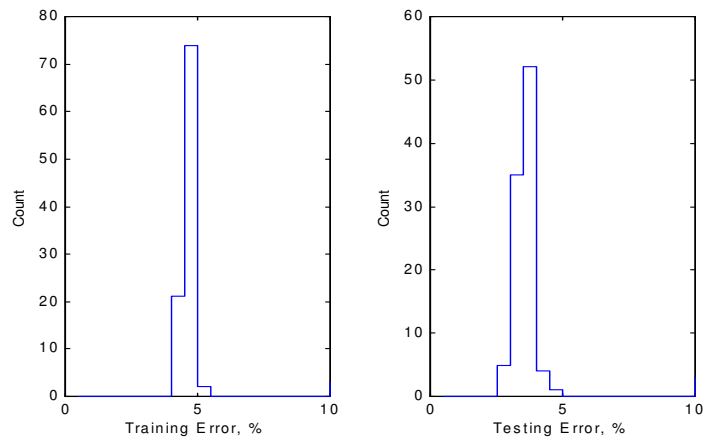

*Figure 3*. A histogram of errors on 100 runs. The error rate ranges from 3.92% to 5.45% on the training set and from 2.73% to 4.96% on the testing set.

At the same time, the training and testing error rates were also varied during these runs. The distribution of these errors is depicted in the histogram, Fig. 3. The minimal training and testing error rates are equal to 3.92% and 2.73%, respectively.

**6. Comparison to the Feed-Forward Neural Networks**

We used the feed-forward neural networks (FNNs) with one hidden layer and one output neuron. The number of hidden neurons we varied from 2 to 8 neurons. All neurons implemented a standard sigmoid transfer function.

To remove the contribution of the correlated inputs and improve the results we have applied a standard preprocessing technique of Principle Component Analysis (PCA). To find the best performance of the PCA, we have given the different fractions *fr* of the total variation in the training dataset so that the number of main components was varied from 72 to 2.

For training the FNN, we have exploited the fast Levenberg-Marquardt (LM) algorithm, provided by MATLAB. In order to prevent the network from over-fitting in this algorithm, we have used the early stopping rule. For this reason, non-intersecting fractions of the dataset have been preserved for training, testing and validating the FNN.

For each given variant of the PCA and the number of hidden neurons, we have trained the FNN 100 times with randomly initiated weights. For each trained FNN we also calculated its error on the testing set.

During these experiments we found that the FNN with 4 hidden neurons and 11 main components that the PCA with $fr = 0.02$ produced performs best on the training set. The training and testing error rates are equal to 2.97% and 5.54%.

We can see that a trained FNN makes more errors on the testing dataset than an ECNN. Hence we can conclude that a standard neural network technique, described above, can provide the over-complicated classifiers.

In our experiments, the error of the 100 trained FNNs varied from 2.77% to 8.71%. The distributions of the training and testing error rates of the FNN are depicted in histograms Fig. 4.

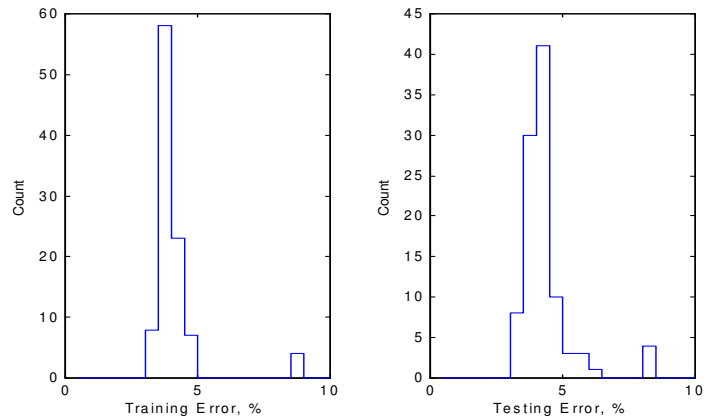

*Figure 4*. A histogram of errors on 100 runs. The error ranges from 2.77% to 8.71% on the training set and from 3.06% to 7.93% on the testing set.

## 7. Discussion

We present in Table 1 the training and testing errors of the best ECNN, FNN, and FNN* on the EEG data. Note that the FNN* uses a neural network structure discovered by the ECNN.

9—*Table 1.* The error rates of the trained ECNN, FNN and FNN*.

|      | Error rates, % | |
|------|-------|------|
|      | Train | Test |
| ECNN | 3.92  | 3.31 |
| FNN  | 2.97  | 5.54 |
| FNN* | 3.30  | 2.98 |

Fist of all we can see that the ECNN outperforms the fully connected FNN on the same testing dataset: minimal errors of ECNN and FNN are equal to 3.31% and 5.54%, respectively. That is, the ECNN algorithm that we developed is able to better prevent over-fitting of neural networks than a standard neural network technique.

Secondly, the suggested algorithm has selected from the initial 72 features the 4 relevant features and found 3 hidden neurons. That is, the algorithm automatically discovers a neural network structure appropriate to the training dataset.

Thirdly, the use of the discovered structure in the fully connected FNN* reduces the testing error to 2.98%. That is, the use of the algorithm as the preprocessing technique is more effective than a standard neural network technique based on PCA.

All computational experiments described here have been carried out with MATLAB 5. In these experiments the learning time of the ECNNs did not exceed the time spent by the fast LM algorithm. It depends on the learning parameters $\chi$ and $\Delta$ in the equation (5) and (6). Nevertheless, of the 2422 training examples, the ECNN took no longer than 14 seconds to learn.

## 8. Conclusion

We have developed a new algorithm for training cascade neural networks to avoid noisy and redundant features during learning. A network starts to learn with a small number of inputs, and then it adds new inputs as well as new hidden neurons during the learning process, thus evolving to a larger network structure. As a result, the trained network has a nearly optimal architecture.

The training algorithm was applied to a real-world problem related to classification of normal segments and artifacts in the EEG recordings. The EEG segments are characterized by several noisy and irrelevant features. The artifacts in the EEG recordings of two patients were visually labeled by EEG-viewer. The ECNN has successfully learned to automatically classify the EEG segments.
The ECNN, trained on the EEG segments, has correctly classified 96.69% of the testing segments. A standard feed-forward network using a PCA preprocessing applied to the same datasets has provided 94.46% of correct classifications. Thus, the ECNN algorithm applied to the EEG problem has performed slightly better than a standard neural network technique. We conclude that the new algorithm can be effectively used to train cascade neural networks applied to real-world problems, which are characterized by many features.

**Acknowledgments**

The author is grateful to Frank Pasemann from TheoriLabor for fruitful and enlightening discussions and to Joachim Frenzel and Burghart Scheidt, Pediatric Clinic of the University Jena, for making available their EEG recordings.

**References**

1. Fahlman, S.E. and Lebiere, C.: The cascade-correlation learning architecture, In: *Advances in Neural Information Processing Systems*, Vol. 2, Morgan-Kauffman, Los Altos, 1990.
2. Smieja, F.J.: Neural network constructive algorithms: Trading generalization for learning efficiency, *Circuits, Systems and Signal Processing* **12** (1993), 331-374.
3. Cios, K.J. and Liu, N.: A machine method for generation of neural network architecture: A continues ID3 algorithm, *IEEE Transactions on Neural Networks* **3** (1992), 280-291.
4. Jang, J. and Honovar, V.: Feature subset selection using a genetic algorithm, In: *Proceedings of the Genetic Programming*, Stanford, 1997.
5. Tetko, I.V. and Villa, A.E.: An enhancement of generalization ability in cascade-correlation algorithm by avoidance of overfitting problem, *Neural Processing Letters* **1** (1997), 43-50.
6. Tetko, I.V. et al: Variable selection in the cascade-correlation learning architecture, In: *Proceedings of 12th European Symposium on Quantitative Structure-Activity Relationships*, Copenhagen, Denmark, 1998.
7. Madala, H.R. and Ivakhnenko, A.G.: *Inductive Learning Algorithms for Complex Systems Modeling*, CRC Press Inc., Boca Raton, 1994.
8. Schetinin, V.: Polynomial neural networks for classifying EEG signals, In: *Proceedings of NIMIA-SC2001 NATO Advanced Study Institute on Neural Networks for Instrumentation, Measurement, and Related Industrial Applications*, Crema, Italy, 2001.
9. Galicki, M., Witte, H., Dörschel, J., Doering, A., Eiselt, M. and Grießbach, G.: Common optimization of adaptive preprocessing units and a neural network during the learning period: Application in EEG pattern recognition, *Neural Networks* **10** (1997), 1153-1163.
10. Riddington, E., Ifeachor, E., Allen, E., Hudson, N. and Mapps, D.: A fuzzy expert system for EEG interpretation, In: *Proceedings of Int. Conference on Neural Networks and Expert Systems in Medicine and Healthcare*, University of Plymouth, 1994.
11. Breidbach, O., Holthausen, K., Scheidt, B. and Frenzel, J.: Analysis of EEG data room in sudden infant death risk patients, *Theory Bioscience* **117** (1998), 377-392.